\pdfoutput=1

\documentclass[11pt]{article}
\usepackage[table,dvipsnames]{xcolor}

\usepackage[final]{acl}

\usepackage{amsmath}
\usepackage{times}
\usepackage{latexsym}
\usepackage{todonotes}
\usepackage{url}
\usepackage[T1]{fontenc}

\usepackage[utf8]{inputenc}

\usepackage{microtype}

\usepackage{inconsolata}

\usepackage{graphicx}

\usepackage{amsmath}
\usepackage{svg}
\usepackage{tabularx}
\usepackage{csquotes}
\usepackage{xspace}
\usepackage{xcolor}
\newcommand{\middletilde}{\raisebox{0.5ex}{\texttildelow}}
\newcommand{\bookcorefpipeline}{BookCoref Pipeline}
\newcommand{\bookcoref}{\textsc{BookCoref}\xspace}
\newcommand{\bookcorefsilver}{\textsc{BookCoref\textsubscript{\textit{silver}}}\xspace}
\newcommand{\bookcorefgold}{\textsc{BookCoref\textsubscript{\textit{gold}}}\xspace}
\newcommand{\bookcorefgoldwindows}{\textsc{Split-BookCoref}\textsubscript{\textit{gold}}\xspace}
\newcommand{\maverickxl}{Maverick\textsubscript{\textit{xl}}\xspace}
\newcommand{\longdoc}{Longdoc\xspace}
\newcommand{\dualcache}{Dual cache}

\newcommand{\bookcorefgoldwindoweval}{\textsc{BookCoref\textsubscript{\textit{gold+window}}}}
\newcommand{\ceafe}{CEAF\textsubscript{\(\phi_4\)}}
\newcommand{\bcubed}{B\textsuperscript{3}}
\NewDocumentCommand{\corefRed}{m m}{{\textcolor{Red}{\textbf{[}}#1\textcolor{Red}{\textbf{]}\textsubscript{#2}}}}
\NewDocumentCommand{\corefGreen}{m m}{{\textcolor{Green}{\textbf{[}}#1\textcolor{Green}{\textbf{]}\textsubscript{#2}}}}
\NewDocumentCommand{\corefBrown}{m m}{{\textcolor{Brown}{\textbf{[}}#1\textcolor{Brown}{\textbf{]}\textsubscript{#2}}}}
\NewDocumentCommand{\corefBlue}{m m}{{\textcolor{Blue}{\textbf{[}}#1\textcolor{Blue}{\textbf{]}\textsubscript{#2}}}}
\NewDocumentCommand{\corefPurple}{m m}{{\textcolor{Purple}{\textbf{[}}#1\textcolor{Purple}{\textbf{]}\textsubscript{#2}}}}

\NewDocumentCommand{\corefOlive}{m m}{{\textcolor{olive}{\textbf{[}}#1\textcolor{olive}{\textbf{]}\textsubscript{#2}}}}

\title{\textsc{BookCoref}: Coreference Resolution at Book Scale}

\author{Giuliano Martinelli\Thanks{Equal contribution.}, Tommaso Bonomo\footnotemark[1], Pere-Lluís Huguet Cabot,\\
 \normalfont{and} {\bf Roberto Navigli}\\
Sapienza NLP Group, Sapienza University of Rome\\
\texttt{\{martinelli, bonomo, huguetcabot, navigli\}@diag.uniroma1.it}}

\begin{document}
\maketitle

\begin{abstract}

Coreference Resolution systems are typically evaluated on benchmarks containing small- to medium-scale documents.
When it comes to evaluating long texts, however, existing benchmarks, such as LitBank, remain limited in length and do not adequately assess system capabilities at the book scale, i.e., when co-referring mentions span hundreds of thousands of tokens.
To fill this gap, we first put forward a novel automatic pipeline that produces high-quality Coreference Resolution annotations on full narrative texts. 
Then, we adopt this pipeline to create the first book-scale coreference benchmark, \bookcoref, with an average document length of more than 200,000 tokens.
We carry out a series of experiments showing the robustness of our automatic procedure and demonstrating the value of our resource, which enables current long-document coreference systems to gain up to +20 CoNLL-F1 points when evaluated on full books.
Moreover, we report on the new challenges introduced by this unprecedented book-scale setting, highlighting that current models fail to deliver the same performance they achieve on smaller documents.
We release our data and code to encourage research and development of new book-scale Coreference Resolution systems at {\small\url{https://github.com/sapienzanlp/bookcoref}}.

\end{abstract} 

\section{Introduction}

\begin{figure}[ht!]
    \centering
    \includegraphics[width=0.99\linewidth]{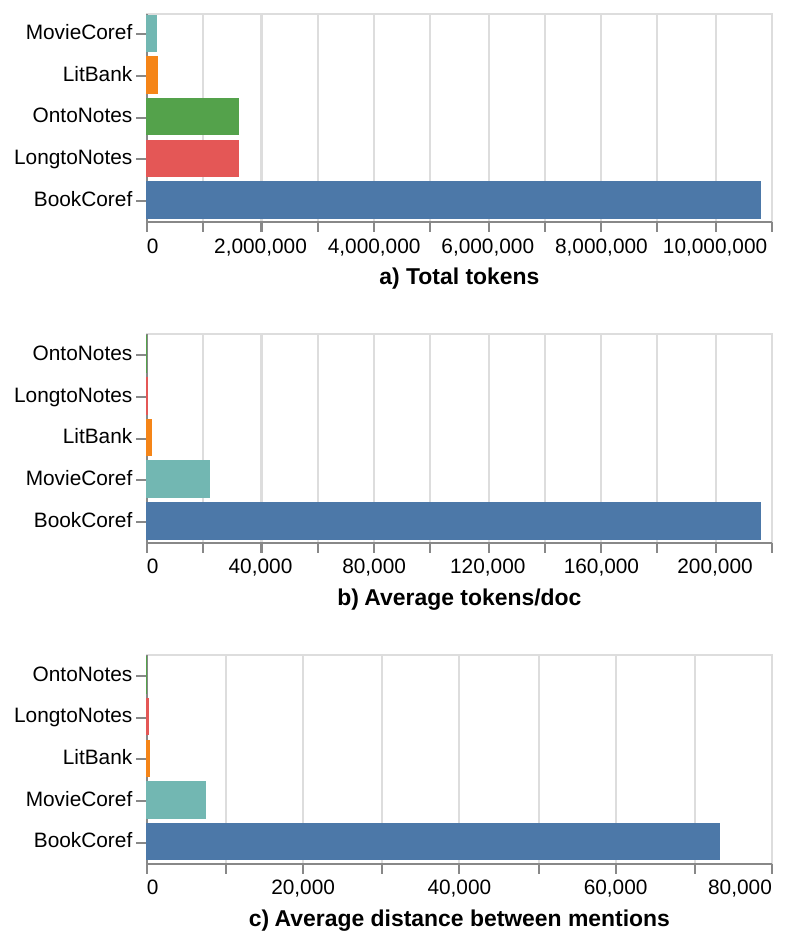}
    \caption{Comparison between \bookcoref\ and current long-document resources, measuring: a) the total number of tokens, b) the average number of tokens per document, and c) the micro-average pairwise distance between all co-referring mentions.}
    \label{fig:introduction-comparison-figure}
\end{figure}

Coreference Resolution (CR) aims to identify and group mentions that refer to the same entity \cite{karttunen-1969-discourse-referents}. 
Co-referring mentions are usually found across sentences and can be located far apart within the same document.
For this reason, as the document length increases, so does the difficulty of manually annotating a corpus with coreference relations \cite{roesiger-etal-2018-towards}.
This is because humans typically resolve coreference relations incrementally \citep{altmann-etal-1988-interaction}, and therefore -- when annotating a coreference mention -- they must rely on previously annotated entities and take the preceding context into account \cite{roesiger-etal-2018-towards}.
This labor-intensive process results in a higher cost of human annotations of long-form texts. As a result, current benchmarks ease the annotation process by drawing samples from short- or medium-sized textual genres (i.e., news, broadcasts, and magazines), or by artificially shortening documents.
For instance, the two most widely used English coreference benchmarks, OntoNotes \cite{pradhan-etal-2013-towards} and LitBank \cite{bamman-etal-2020-annotated}, either split documents into smaller chunks or truncate them to the first 2,000 tokens.
For this reason, most CR systems are optimized for shorter texts and struggle to handle longer inputs effectively.

Several recent works, such as LongtoNotes \cite{shridhar-etal-2023-longtonotes} and MovieCoref \citep{baruah-etal-2021-annotation, baruah-narayanan-2023-character}, attempt to address this gap by introducing manually-annotated long-document CR resources. 
However, LongtoNotes documents remain limited in length (less than 700 tokens per document), while MovieCoref contains only a small number of full-length screenplays, a niche narrative genre with an atypical textual structure.
We argue that the current lack of book-scale benchmarks leaves the many challenges this setting poses -- such as efficient processing~\cite{toshniwal-etal-2020-learning, toshniwal-etal-2021-generalization}, consistency~\cite{guo-etal-2023-dual}, and evaluation~\cite{durontejedor2023evaluatecoreferenceliterarytexts} of Coreference Resolution in long documents -- largely unexplored.

To address this limitation, we introduce a highly reliable automatic pipeline to annotate long documents, which we then apply to full-length literary works to create the first book-scale coreference benchmark, \bookcoref.
Inspired by recent studies that suggest focusing on a smaller yet relevant set of entities \cite{baruah-etal-2021-annotation, guo-etal-2023-dual, manikantan-etal-2024-major}, our annotation involves only the characters that appear in a book, as they are the main agents of fictional stories \citep{bamman-etal-2013-learning, roesiger-etal-2018-towards, labatut-etal-2019-extraction}.
As shown in Figure~\ref{fig:introduction-comparison-figure}, \bookcoref\ presents unprecedented long-document characteristics, enabling the study of the coreference phenomena in full narrative books, which was previously impossible. To summarize, in this work:
\begin{itemize}
    \item We put forward the \bookcorefpipeline, a novel procedure that enables Coreference Resolution annotation of full documents, and extensively validate its effectiveness.
    \item We introduce \bookcoref, a new book-scale Coreference Resolution dataset with a manually-annotated split, \bookcorefgold, to train and evaluate systems on fully annotated books.\footnote{We release \bookcoref\ on HuggingFace: {\small\url{https://huggingface.co/datasets/sapienzanlp/bookcoref}}.}
    \item We benchmark state-of-the-art Coreference Resolution systems, reporting on the open challenges of this new book-scale setting.
\end{itemize}

\section{Related Work} \label{section:related-work}
We now review established resources and approaches that deal with long-document Coreference Resolution.
In Section \ref{section:related-work-benchmarks}, we delve into the details of the available datasets, underscoring the current lack of training and evaluation resources at the book scale.
Subsequently, in Section \ref{section:related-work-model}, we survey recent state-of-the-art CR systems, highlighting that many are not well suited for processing long documents.

\subsection{Long-document Benchmarks} \label{section:related-work-benchmarks}
Existing English Coreference Resolution benchmarks typically feature short- to medium-sized documents. 
Among these, OntoNotes~\cite{pradhan-etal-2013-towards} is the most widely used dataset, comprising documents from various genres, such as news, broadcast, and magazines, with an average length of 467 tokens. 
This brevity arises from splitting source documents into smaller partitions to simplify annotation. 
\citet{shridhar-etal-2023-longtonotes} manually merge coreference clusters across partitions to construct full-document annotations (LongtoNotes), but, due to the short nature of the source genre, the average length only increases to 679 tokens.

WikiCoref~\cite{ghaddar-langlais-2016-wikicoref}, instead, explores the encyclopedic genre, proposing a 37-document evaluation set of Wikipedia pages with an average length of 1,995 tokens. 
Similarly, \citet{bamman-etal-2020-annotated} introduce LitBank, an annotated benchmark of 100 book samples from the literature genre, where documents are long, and relatively few major entities play a central role.
However, while LitBank has become the long-document standard for CR, it truncates book samples to 2,000 tokens, failing to capture coreference relations that develop across entire books.

Recent work explores the Coreference Resolution task in longer contexts.
MovieCoref~\cite{baruah-etal-2021-annotation, baruah-narayanan-2023-character} consists of 6 full-sized screenplays and 3 excerpts, each manually annotated with coreference relations, reaching an average text length of 20,000 tokens.
However, the number of annotated documents is relatively small, and their coreference annotations depend heavily on the underlying screenplay structure, which is very different from the free-form text found in other coreference datasets.
Finally, \citet{guo-etal-2023-dual} manually annotate \textit{Animal Farm} by George Orwell as a benchmark to test the capabilities of CR systems when applied to longer text.
Notably, both the annotation guidelines of MovieCoref and \textit{Animal Farm} focus exclusively on characters, reflecting their central role in modern narrative analysis \cite{piper-etal-2021-narrative, manikantan-etal-2024-major}.
In \bookcoref, we follow the same design choice and build a large-scale dataset, which we then use to train and evaluate CR capabilities of systems on a diverse set of narrative books.

\subsection{Long-document Coreference Systems}  \label{section:related-work-model}
Most recent state-of-the-art solutions for CR are specifically tailored to short- or medium-length documents and are not suited to the book-scale setting.
Generative approaches that formulate CR as a sequence-to-sequence task \cite{bohnet-etal-2023-coreference, zhang-etal-2023-seq2seq} usually require re-generating the entirety of the input text along with the coreference annotations, effectively doubling the context length.
This is impractical for book-scale settings, as generative systems usually rely on very large Transformers, which have fixed input lengths and imply a high computational cost that becomes prohibitive when processing entire books.
The same concerns apply to Large Language Models (LLMs), whose application to CR is still under discussion: current methods for LLM-based CR have yet to reach the performance of supervised models~\citep{le2024are, porada-cheung-2024-solving}.

Discriminative encoder-only models are in general more memory- and time-efficient \cite{otmazgin-etal-2022-f} but suffer from a similar limitation: recent approaches such as LingMess \cite{otmazgin-etal-2023-lingmess} or s2e-coref \cite{kirstain-etal-2021-coreference} are constrained to respect the maximum input length of their underlying Transformer encoder, i.e., LongFormer~\cite{beltagy-etal-2020-longformer}.
On the other hand, some recent encoder-only solutions can handle longer documents:
the current state-of-the-art coreference resolution system, Maverick~\cite{martinelli-etal-2024-maverick}, is built upon DeBERTa-v3~\cite{he-etal-2023-debertav}, which can encode up to \text{25,000} tokens~\cite{he-etal-2021-deberta}.
However, due to its quadratic computational complexity with respect to input length, processing entire books remains impractical, as hardware requirements scale rapidly.

To solve these issues, recent work proposes systems tailored to the processing of long text, employing an incremental formulation.
\longdoc~\cite{toshniwal-etal-2020-learning, toshniwal-etal-2021-generalization} specifically tackles the runtime memory problems that long-document CR causes: it incrementally builds entity coreference clusters and learns to ``forget'' entities using a global cache of the most recent entities predicted.
More recently, \dualcache~\cite{guo-etal-2023-dual} proposes modifying the \longdoc\ architecture by introducing a secondary cache that globally takes into account the least-frequently used mentions.
As noted in Section \ref{section:related-work-benchmarks}, the researchers measure their improvements by annotating a full book, Orwell's \textit{Animal Farm}.
Interestingly, they show that their formulation outperforms other systems, but only reaches 36\% CoNLL-F1, which is much lower than the average scores on other medium-sized datasets.

This highlights that i) a more complete book-level training and evaluation benchmark is needed to assess system performance on extended contexts, and that ii) current long-document systems cannot be used as automatic annotators, an issue we address through the \bookcorefpipeline.


\section{\bookcoref}



\begin{figure*}[ht!]
    \centering
    \includegraphics[width=0.99\linewidth]{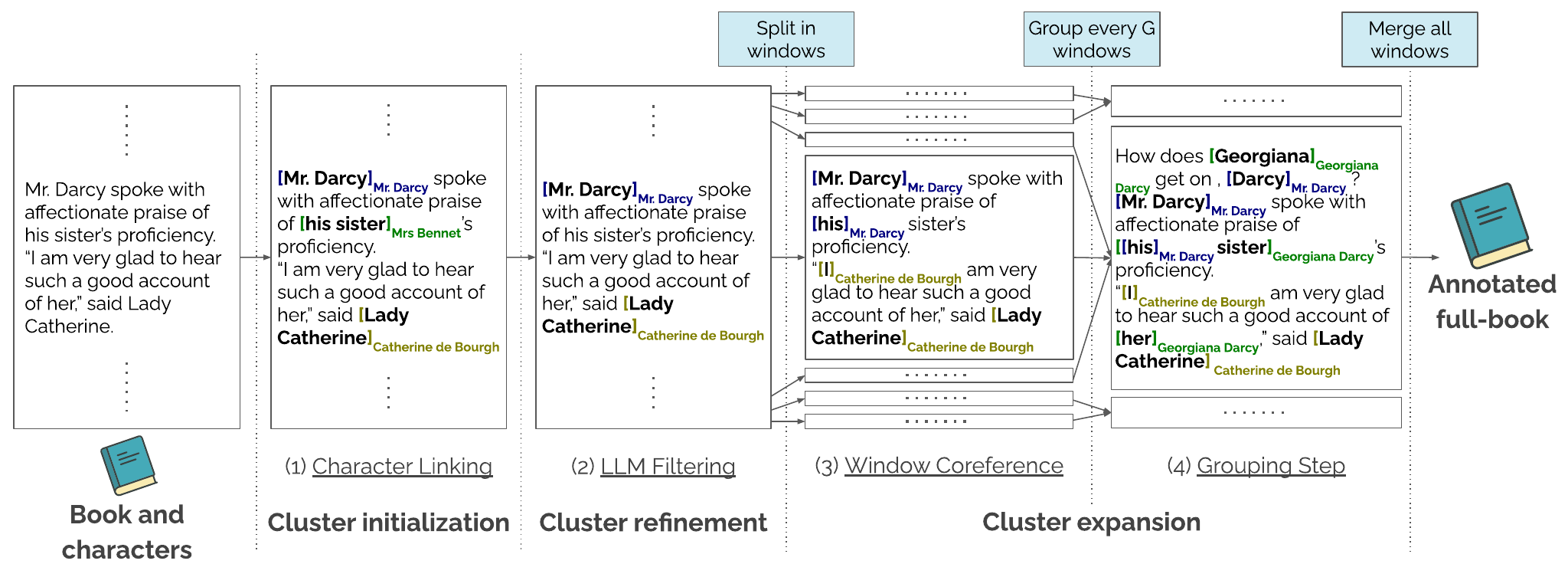}
    \caption{The \bookcorefpipeline\ applied on a sample taken from \textit{Pride and Prejudice}. (1) Link all explicit character mentions via Character Linking. (2) Filter out inconsistent assignments via LLM Filtering. (3) Expand character clusters using a CR model on small windows. (4) Expand character clusters of grouped windows.}
    \label{fig:bookcoref-pipeline}
\end{figure*}

This Section describes the resources and methods we used to create our novel book-level benchmark, \bookcoref.
Specifically, in Section~\ref{subsection:resources} we introduce the underlying resources used to obtain the full texts and the character list of the books we annotate.
Then, in Section~\ref{subsection:pipeline}, we present the \bookcorefpipeline, our automatic procedure for producing high-quality silver data for book-level CR.
Finally, in Sections~\ref{subsection:manual} and \ref{subsection:statistics} we detail our manual annotation process and compare the resulting \bookcoref\ corpus with previous well-established CR benchmarks.

\subsection{Underlying Resources \label{subsection:resources}}
Our automatic pipeline annotates full-text documents starting from their list of characters.
Therefore, to produce \bookcoref\ we leverage three main resources:
i) Project Gutenberg\footnote{{\small\url{https://www.gutenberg.org/}}}, a collection of openly available literary works;
ii) Wikidata, a multilingual knowledge graph containing thousands of books\footnote{As of 2024-12-09, Wikidata contains 75,675 book items.};
iii) LiSCU~\citep{brahman-etal-2021-characters-tell}, a dataset that collects information from online study guides on the characters appearing in English literary works.

We start by collecting a list of characters, authors, and titles of books available in Wikidata and LiSCU. As these resources do not provide full texts, we search Project Gutenberg for complete books that match the corresponding authors and titles. To ensure high quality, we manually remove erroneous matches and automatically exclude books with fewer than five characters.

Formally, after this step, we obtain a set of full-text books \(\mathcal{B}\) where each book \(b \in \mathcal{B}\) is paired with a list of character names \(\mathcal{C}(b) = \{c_1, c_2, \dots, c_n\}\).
Our corpus contains \(|\mathcal{B}| = 53\) books with an average of 27 characters per book, mostly comprising full-length classical novels such as Walter Scott's \textit{Ivanhoe} or Jane Austen's \textit{Pride and Prejudice}.
In Appendix~\ref{appendix:liscu-gutenberg} we provide a more detailed analysis of our corpus and preprocessing step, along with a detailed table of all the books included in \bookcoref.

\subsection{\bookcorefpipeline \label{subsection:pipeline}}
We now introduce the \bookcorefpipeline, our proposed automatic annotation procedure.
Formally, the \bookcorefpipeline\  takes as input a book \(b\) and its set of character names \(\mathcal{C}(b)\) and outputs coreference cluster annotations over the full book for each character name.
Notably, although we apply our pipeline to narrative books, it is genre-agnostic and can be exploited for any long document, provided a set of named entities is available.
Figure~\ref{fig:bookcoref-pipeline} summarizes the steps of our \bookcorefpipeline, namely, Cluster Initialization (Section \ref{subsub:cluster-init}), Cluster Refinement (Section \ref{subsub:cluster-refinement}), and Cluster Expansion (Section \ref{subsub:cluster-expansion}).

\subsubsection{Cluster Initialization \label{subsub:cluster-init}}
In our first step, we initialize character coreference clusters\footnote{By character coreference cluster, we refer to the set of coreferential mentions that correspond to the same character.} by tagging explicit entity mentions (i.e., non-pronoun mentions such as proper nouns or noun phrases).
We extract these mentions through Entity Linking (EL), the task of linking explicit mentions of entities in a text to their corresponding entry in a Knowledge Base (KB).

In our case, the index (or KB) of possible entities changes for each book \(b\) and corresponds to the list of character names \(\mathcal{C}(b) = \{c_1, c_2, \dots, c_n\}\).
Formally, given a character \(c\) from a book \(b \in \mathcal{B}\), we initialize its coreference cluster \(\mathcal{E}^b_c\) with all its explicit mentions extracted through an automatic EL system.
We note as \(\left\{\mathcal{E}^b_c\right\}_{c \in \mathcal{C}(b)} = \left\{\mathcal{E}^b_{c_1}, \mathcal{E}^b_{c_2}, \dots, \mathcal{E}^b_{c_n} \right\}\) the set of all character coreference clusters in a book \(b\). 

The main limitation of current pre-trained EL systems is that they are trained to link named entities to predefined knowledge bases, such as Wikipedia or Wikidata; in our case, instead, we aim to link explicit mentions of characters to their respective names, a task that we define as Character Linking.
To this end, we prepare a small training set for the Character Linking task.
Specifically, we manually amend LitBank~\cite{bamman-etal-2020-annotated} by filtering out all non-explicit character mentions and naming each cluster with character names (full details are provided in Appendix~\ref{appendix:relik-litbank}).
We then use this resource to fine-tune a state-of-the-art EL system \citep[ReLiK]{orlando-etal-2024-relik} for Character Linking in the narrative genre.

We apply this trained system to our collection of books \(\mathcal{B}\), obtaining the aforementioned character coreference clusters \(\left\{\mathcal{E}^b_c\right\}_{c \in \mathcal{C}(b)}\), for each book \(b \in \mathcal{B}\).
As a by-product, our clusters are linked to a specific name, which is crucial to the next steps of our pipeline.

\subsubsection{Cluster Refinement \label{subsub:cluster-refinement}}
We place a strong emphasis on the precision of our initial cluster formation:
false positives (i.e., incorrect mention-character links predicted by our system) can easily propagate through the whole pipeline, whilst false negatives (i.e., correct mention-character links that are not predicted by our system) are easier to recover in our subsequent Cluster Expansion step.
Therefore, we introduce an additional verification step for each mention of a character cluster by prompting an LLM to ascertain whether the mention is accurately linked to the character, based on the surrounding context.

Specifically, we prompt Qwen2 7B Instruct~\citep{yang2024qwen2technicalreport}\footnote{We chose Qwen2 7B due to its permissive Apache 2.0 license.} with the name of the character and the highlighted mention in context, and we constrain the LLM to output either `Yes' or `No' as the answer to our prompt (see Appendix~\ref{appendix:llm} for the exact prompt and further details).
We filter out mentions for which the LLM answers negatively.
The final result of this step is a set of LLM-validated coreferential clusters \(\left\{\hat{\mathcal{E}}^b_c\right\}_{c \in \mathcal{C}(b)}\) for each character \(c\) of all our books \(\mathcal{B}\).

\begin{table*}[]
    {\small
    \begin{tabularx}{\linewidth}{|>{\raggedleft\arraybackslash}X ||c|c|c|c|c|c|c|c||}
    \hline
    Datasets
    & Docs
    & Tok.
    & Ment.
    & Tok./Doc
    & Ment./Doc
    & Chains/Doc
    & Ment./Chain
    & Ment. Dist. \\ 
    \hline
    OntoNotes & 3,493 & ~~1.6M & 194k & ~~~~~~~467 & ~~~~~~~55 & 12.7 &~4.3 & ~~~~~~~140 \\
    LongtoNotes & 2,415 & ~~1.6M & 194k & ~~~~~~~674 & ~~~~~~~80 & 16.7 &~4.9 & ~~~~~~~371 \\
    LitBank & ~~100 & ~~210k & ~~29k & ~~~~2,105 & ~~~~~291 & 79.3 & ~4.1 & ~~~~~~~480\\
    WikiCoref & ~~~~30 & ~~~~59k & ~~~~7k & ~~~~1,996 & ~~~~~233 & 49.4 &~4.5 & ~~~~1,013 \\
    MovieCoref & ~~~~~~9 & ~~201k & ~~26k & ~~22,423 & ~~2,865 & 46.4 &~~89 & ~~~~7,672 \\                             
    \hline
    \bookcorefgold & ~~~~~~3 & ~~229k & ~~24k & ~~76,419 & ~~7,844 & 22.3 & 359 & ~~34,880 \\
    \bookcorefsilver & ~~~~50 & 10.8M & 968k & 216,626 & 19,471 &  27.4 & 870 & ~~73,432 \\
    \hline
    \end{tabularx}%
    }
    \caption{Statistics of \bookcoref\ compared with previous CR  datasets. Columns from left to right: total number of documents, tokens, and mentions; average number of tokens, mentions, and coreference chains per document; micro-average number of mentions per chain; and micro-average pairwise distance between mentions of the same chain.
    We use (k) and (M) to indicate thousands and millions, respectively.
    }
    \label{tab:doc-coref-datasets}
\end{table*}

\subsubsection{Cluster Expansion \label{subsub:cluster-expansion}}

As a result of the previous steps, for each book we have a set of high-precision annotations of coreference clusters containing the explicit mentions of each character that appears in that particular book.

To fully tag our books with complete CR annotations, we expand each coreference cluster by including every missing mention of a given character, including pronouns, noun phrases, and other references not tagged in the previous steps.
We employ an automatic CR system, Maverick~\citep{martinelli-etal-2024-maverick}, because it combines state-of-the-art performance with the ability to complete partial coreferential clusters, which is a prerequisite in order to exploit the annotations obtained in the previous steps.
However, the problem with using Maverick in a long-context setting is that it suffers from the same input length issues that impact other CR models, as discussed in Section~\ref{section:related-work-model}.
Therefore, for each book \(b \in \mathcal{B}\), we apply Maverick to consecutive, non-overlapping windows \(\mathcal{W}(b) = \{w_1, \dots, w_N\}\), with each window \(w_i\) containing a maximum number of 1500 words, the closest setting to Maverick's training conditions.
More precisely, for each window $w_i\in \mathcal{W}(b)$ in book $b$, we derive from the set of coreferential clusters produced in the previous steps \(\left\{\hat{\mathcal{E}}^b_c\right\}_{c \in \mathcal{C}(b)}\) a local set of clusters \(\left\{\hat{\mathcal{E}}^{w_i}_c\right\}_{c \in \mathcal{C}(b)}\) by filtering out all mentions of a cluster that are not contained in window $w_i$.
Following this, we use Maverick on window \(w_i\) to expand each character cluster \(\hat{\mathcal{E}}^{w_i}_c\) to all of its coreferential mentions, including pronouns, noun phrases and other references, obtaining a local and complete set of clusters \(\left\{\mathcal{M}^{w_i}_c\right\}_{c \in \mathcal{C}(b)}\) for a window \(w_i\) of a book \(b\).

To obtain a full-book annotation for \(b\), we merge the annotations from all the windows \(w_i \in \mathcal{W}(b)\).
This can be done by taking the union of every cluster linked to the same character \(c\) across all windows: 
\(\mathcal{M}^b_c = \bigcup^{N}_{i=1} \mathcal{M}^{w_i}_c\).
We leverage the fact that local coreference clusters are each linked to the same global set of character names, as discussed in the previous section.
Although merging all windows $w_i \in \mathcal{W}(b)$ results in a precise annotation of book $b$, this approach does present a particular edge case that affects the recall of character mentions:
when a window $w_i$ lacks explicit mentions of a character $c \in \mathcal{C}(b)$, our Cluster Expansion step becomes ineffective, as there are no mentions from which to expand and form the corresponding coreferential cluster.
This is outlined in Figure~\ref{fig:bookcoref-pipeline}, where the mention ``his sister'' is not linked to the character \textit{Georgiana Darcy}.

To mitigate this limitation, we propose an intermediate grouping step in which we merge $G$ consecutive windows into a single grouped window\footnote{By performing a statistical analysis on our corpus, we found that $G=10$ results in an optimal intermediate length between the size of windows and full books.}, and run a second expansion step on a context that is larger and contains more characters.
We therefore introduce \maverickxl\  (Section \ref{subsection:systems}), an adaptation of Maverick that processes longer documents by encoding them in smaller splits, and use it to perform this second Cluster Expansion step.
Formally, we define non-overlapping, grouped windows $gw_j \in \{gw_1, \dots, gw_M\}$, with $M = \left\lceil|\mathcal{W}(b)|\; /\;G \right\rceil $, as the union of $G$ consecutive windows, \(gw_j = \bigcup_{i=G \cdot (j-1) + 1}^{\text{min}(G \cdot j,N)} w_i\), where $w _i \in \mathcal{W}(b)$. 
For each \(gw_j\), we define its set of character clusters as the union of the local coreference clusters of the grouped windows: \(\left\{\mathcal{M}^{gw_j}_c\right\}_{c \in \mathcal{C}(b)} = \left\{\bigcup^{\text{min}(G \cdot j,N)}_{i=G \cdot (j - 1) + 1} \mathcal{M}^{w_i}_c\right\}_{c \in \mathcal{C}(b)}\).
This grouping step expands the character clusters by considering the broader context provided by the grouped windows.
We apply \maverickxl to each grouped window $gw_j$ starting from $\left\{\mathcal{M}^{gw_j}_{c}\right\}_{c \in \mathcal{C}(b)}$ and we obtain a local set of coreference resolution annotations $\left\{\mathcal{F}^{gw_j}_{c}\right\}_{c \in C(b)}$.
The benefits of this step can be seen clearly in Figure~\ref{fig:bookcoref-pipeline}, where the mention ``his sister'' is correctly linked to the character \textit{Georgiana Darcy}.
Finally, for each character cluster, we take the union across these grouped windows \(gw_j\), obtaining coreference annotations $\left\{\mathcal{F}^b_{c}\right\}_{c \in C(b)}$
for each book, which altogether represent the final coreference annotation of \bookcorefsilver.




\subsection{Manually-annotated Test Set \label{subsection:manual}}
To enable a complete and rigorous evaluation of coreference resolution models at book scale, we put forward a manually annotated benchmark, \bookcorefgold.
We include the annotation provided by \citet{guo-etal-2024-dual} for \textit{Animal Farm} by George Orwell and two new, fully-annotated books, i.e., Herman Hesse's \textit{Siddhartha} and Jane Austen's \textit{Pride and Prejudice}.
This latter is a particularly challenging benchmark as previous work has shown that its high density of characters and pronominal expressions often leads CR systems to produce erroneous links~\citep{vala-etal-2015-mr}.

Our annotation guidelines mirror the ones of \textit{Animal Farm} and are further explained in Appendix~\ref{appendix:annotation-guidelines}.
Each annotator is presented with the full text of the book and the corresponding list of characters, and is tasked with tagging any mention of the characters by means of an annotation tool.\footnote{\small\url{https://github.com/nilsreiter/CorefAnnotator}}
The annotation was conducted by three expert CR annotators (authors of this paper) for around 120 hours, covering a total of 194,280 words. 
Following recent literature, the agreement was computed on a 2,000-word subset using traditional coreference metrics, achieving a 96.1 inter-annotator MUC score -- comparable to LitBank (95.5) and higher than OntoNotes (83.0). Notably, restricting annotations to characters contributes to a high agreement.

\begin{table*}[]
    \centering
{
{\small
\begin{tabularx}{0.9\linewidth}{|>{\arraybackslash}X||c|c|c|| c| c| c| c||}
\hline 

& \multicolumn{3}{c|}{Character Linking} & \multicolumn{4}{c|}{Coreference} \\
\hline

& P & R & F1 & MUC & \bcubed & \ceafe & CoNLL \\ 
\hline

\rowcolor{gray!20} \multicolumn{8}{|c|}{Cluster Initialization} \\

Pattern matching & 95.6 & 18.9 & 29.2 & 14.7 & ~~4.7 & 34.5 & 17.9\\
Character Linking (CL) & 88.6 & 30.9 & 44.5 & 39.3 & 12.7 & 50.5 & 34.2 \\
\hline

\rowcolor{gray!20}\multicolumn{8}{|c||}{Cluster Refinement} \\
CL + LLM filtering & 93.8 & 29.8 & 43.5 & 37.5 & 50.9 & 12.9 & 33.9  \\
\hline

\rowcolor{gray!20} \multicolumn{8}{|c||}{Cluster Expansion} \\
CL + Window Coreference & 85.7 & 75.3 & 80.2 & 85.7  & 62.9  & 65.3  & 71.3 \\
\quad + Grouping step & 78.7 & 87.8 & 83.0 & 91.0 & 65.2 & 67.8  & 74.7\\
\hline
CL +  LLM filtering + Window Coreference & 90.3 & 80.4 & 84.7 & 85.6  & 67.0  & 78.6  & 77.7 \\
\quad + Grouping step & 83.2 & 89.8 & 86.3 & 93.3  & 70.8 & 77.5  & 80.5 \\
\hline 

\end{tabularx}%
}
}

\caption{Evaluation of our \bookcorefpipeline\ on \bookcorefgold. We report precision, recall and F1-score for Character Linking and measure MUC, \bcubed, \ceafe\ and CoNLL-F1 as coreference metrics.}
\label{tab:pipeline-ablation}
\end{table*}

\subsection{Statistics \label{subsection:statistics}}
In Table~\ref{tab:doc-coref-datasets} we report the statistics of our two new resources, \bookcorefsilver\ and \bookcorefgold, compared to well-established long-document benchmarks.
Some distinctly book-scale characteristics emerge, such as very long documents, a low number of highly populated chains for each document, and a high average pairwise distance between mentions of the same chain.\footnote{The average pairwise distance is calculated as the micro-average of all the distances, measured in number of tokens, between any two mentions that are part of the same chain.}
\bookcorefsilver\ is both large-scale, with 6.75 times the number of tagged tokens and 5 times the amount of mentions compared to OntoNotes, and book-level, with mentions linked across unprecedented distances.
Despite containing only three manually-annotated books, we note that \bookcorefgold\ contains 229k tokens, \middletilde10 times more than the test splits of well-established long-document benchmarks, LitBank (\middletilde21k) and MovieCoref (\middletilde29k).

\subsection{\bookcorefpipeline\ Evaluation \label{subsection:pipeline-eval}}

We measure the performance of each step of our \bookcorefpipeline\ on our manually curated subset of books, \bookcorefgold.
Table~\ref{tab:pipeline-ablation} reports traditional CR metrics and standard metrics for Character Linking.
In particular, we measure MUC \cite{vilain-etal-1995-model}, \bcubed\ \cite{bagga-baldwin-1998-entity-based}, \ceafe\ \cite{luo-2005-coreference} and their average value, CoNLL-F1, for Coreference Resolution, and precision, recall and F1-score for Character Linking.
We evaluate our initial Cluster Initialization step performance, comparing our approach to a simple Pattern Matching (PM) baseline, where character clusters are initialized by selecting mentions that match the character name.
Our Character Linking method obtains high precision scores and ensures a higher recall of character mentions compared to PM, resulting in an increase of +15.3 in F1-score.
Applying LLM-based filtering on top of our Character Linking outputs increases the character precision by an average of +5.2 points, as it removes several wrongly-predicted mentions.
Keeping these mentions would result in a propagation of annotation errors in the following Cluster Expansion step: in Appendix Table~\ref{table:error-analysis} we present a qualitative analysis of the impact of this step.
Regarding the final Cluster Expansion step, we show that the intermediate grouping strategy improves coreference-specific metrics, reaching 80.5 CoNLL-F1 score, and therefore we adopt it in the final \bookcorefpipeline.
Finally, we highlight that our silver-annotation pipeline achieves an MUC score of 93.3, comparable to the inter-annotator MUC agreement between our human annotators (96.1) and between the annotators of other datasets (LitBank 95.5, OntoNotes 83.0, cf. Section~\ref{subsection:manual}).

\begin{table*}[t!]
    \centering
    {\small

    \begin{tabularx}{\textwidth}{
        |l|||
            >{\centering\arraybackslash}X| >{\centering\arraybackslash}X| >{\centering\arraybackslash}X ||
            c| c| c|| c|||
            c| c| c|| c|||
    }
               
    \hline
    
    & \multicolumn{7}{c|||}{\bookcorefgold} 
    & \multicolumn{4}{c|||}{\bookcorefgoldwindows} 
               
    \\
    \hline
    Models
    & Ani.
    & P.\&P.
    & Siddh.
    & \textsc{MUC}
    & \textsc{B$^{3}$}
    & \textsc{C\textsubscript{\(\phi_4\)}}
    & \textsc{CoNLL}
    
    & MUC
    & \bcubed
    & \textsc{C\textsubscript{\(\phi_4\)}}
    & \textsc{CoNLL}

    \\ 
    \hline
    \rowcolor{gray!20}
    \multicolumn{12}{|c|||}{Off-the-shelf}
    \\

    BookNLP & \underline{41.1} & 41.2 & 45.1 & \underline{83.1} & 40.9 & 2.4 & 42.2 & 81.1 & 52.3 & 18.7 & 50.6\\
    \longdoc  & 39.1 & \underline{48.9} & 44.8 & 79.9 & \underline{52.1} & \underline{7.9} & \underline{46.6} & 80.7 & 63.8 & 39.0  & 61.2\\
    \dualcache & 36.7 & 42.2 & \underline{50.9} & 82.3 & 41.0 & 4.2 & 42.5 & 82.9 & 67.8 & 43.9 & 64.8 \\
    \maverickxl & 29.1 & 39.2 & 50.8 & 81.2  & 35.6 & 6.1 & 41.2 & \underline{83.8} & \underline{69.5} & \underline{46.1}  & \underline{66.5} \\
    \hline
    \rowcolor{gray!20}
    \multicolumn{12}{|c|||}{Fine-tuned on \bookcorefsilver}
    \\
    \longdoc & \textbf{67.5} & \textbf{63.9} & \textbf{74.0} & 93.5 & \textbf{62.4} & \textbf{45.3} & \textbf{67.0} & 91.2 & 74.9 & 65.7 & 77.1 \\
    
    \dualcache & 52.6 & 47.9 & 68.9 & 92.5 & 48.0 & 16.9 & 52.5 & 91.2 & 74.5 & 66.2 & 77.3 \\
    
    \maverickxl & 62.7  & 57.5 & 68.0 & \textbf{94.3}  & 55.3  & 33.4 & 61.0 & \textbf{92.7} & \textbf{82.2} & \textbf{71.9}  & \textbf{82.2} \\
    
    \hline
    
    \end{tabularx}%
    }
\caption{Comparison between off-the-shelf models and systems trained on \bookcorefsilver, tested on \bookcorefgold\ and \bookcorefgoldwindows. For each system, we report F1 measures of MUC, \bcubed\ and \ceafe\ (noted as C\textsubscript{\(\phi_4\)}), and use their average, CoNLL-F1, as the main evaluation criteria. The first three columns detail avg. CoNLL-F1 scores on the three different books of our test-set, namely \textit{Animal Farm}, \textit{Pride and Prejudice} and \textit{Siddharta}. We highlight in \textbf{bold} the best measures of trained models, and \underline{underline} best off-the-shelf results.}
\label{tab:result}
\end{table*}

\section{Experimental Setup}
In this Section, we report the experimental setup we developed to empirically analyze the training and testing of current systems on \bookcoref.

\subsection{Experiments Design \label{subsection:exp-design}}

We benchmark automatic coreference systems on the proposed \bookcoref\ dataset.
Specifically, we test the models both off-the-shelf, i.e., loading their pre-trained configurations, and after training them on \bookcorefsilver.
We measure CR performance in two specific settings: 
i) on \bookcorefgold, in which models are evaluated on book-scale coreference by taking in input full books; ii) on \bookcorefgoldwindows, in which models are evaluated on medium-sized texts that result from splitting \bookcorefgold\ into independent windows of 1500 tokens.
The two settings allow us to evaluate the ability of current CR systems to overcome the intrinsic difficulty of processing book-scale texts in comparison with medium-sized ones.

\subsection{Comparison Systems \label{subsection:systems}}
As discussed in Section~\ref{section:related-work-model}, many available coreference systems cannot be applied to the book-scale setting.
Different modeling approaches (LLMs, seq-to-seq, encoder-only) present significant limitations, which we describe in detail in Appendix~\ref{appendix:limitations}.
Therefore, in our experiments, we benchmark available solutions for long documents and adapt a state-of-the-art coreference system to the book-scale setting. 
Among long-document systems, we benchmark the widely-adopted BookNLP library\footnote{\small\url{https://github.com/booknlp/booknlp}}, a BERT-based model that is trained on LitBank.
We also include \longdoc~\citep{toshniwal-etal-2020-learning, toshniwal-etal-2021-generalization} and \dualcache~\citep{guo-etal-2024-dual}, two systems that were specifically designed to handle long texts through an incremental formulation of CR.

We compare the results of these systems with Maverick~\cite{martinelli-etal-2024-maverick}, a non-incremental architecture that currently achieves state-of-the-art results on OntoNotes.
To use Maverick on our book-scale setting, we adapt its architecture at inference time: 
instead of encoding documents in their entirety, we perform mention extraction on smaller windows of arbitrary length \(L\).
We then carry out the subsequent clustering steps on all the mentions of the books, drastically reducing the memory requirements that would be involved in encoding the whole book.
We name this model \maverickxl, and we set \(L = 4000\), the largest value that allows us to run inference on \bookcorefgold\ on our academic-budget setup, i.e., a single NVIDIA RTX-4090 with 24GB of VRAM.
Notably, Maverick and \maverickxl\ are equivalent when testing the model on \bookcorefgoldwindows, where the input text is shorter than 4000 tokens.

With the exception of BookNLP\footnote{BookNLP's codebase does not allow training on new datasets.}, we train our comparison systems on \bookcorefsilver. 
We split documents and respective coreference clusters into separate windows of a maximum length of \(1500\) tokens. 
This length both complies with the maximum input length specified by the adopted Transformer encoders, and also enables us to train all the comparison systems on our hardware avoiding the huge memory cost of training on full books\footnote{For reference, training Maverick on a single book of 300k tokens requires more than 1200 GB of VRAM.}.

We reserve Appendix~\ref{appendix:training} for further details on the comparison systems and our training setup.

\section{Results \label{section:results}}
In Table \ref{tab:result}, we benchmark current models on \bookcorefgold\ and on \bookcorefgoldwindows.

\paragraph{\bookcorefgold}
Off-the-shelf models all achieve a CoNLL-F1 score above 40 points, with \longdoc\ obtaining the best performance of 46.6.
Notably, in this setting, the popular BookNLP performs similarly to other neural counterparts.

As expected, training our comparison systems on the \bookcorefsilver\ data consistently improves their performance.
In this setting, \longdoc\ is still the best-performing model, scoring 67.0 CoNLL-F1 points (+20.4 compared to the off-the-shelf version), demonstrating the superiority of its incremental formulation.
Interestingly, \dualcache, which was previously considered the best choice for long-document CR, achieves low scores after fine-tuning, even when compared with \maverickxl, a model that is not specifically tailored for long-document processing.

Our results demonstrate that none of the available off-the-shelf systems could have been considered a good candidate for the annotation of our silver corpus. 
In fact, the proposed \bookcorefpipeline\ obtains 80.5 CoNLL-F1 score on \bookcorefgold\ (Table \ref{tab:pipeline-ablation}), +33.9 compared to the best off-the-shelf model.

Notably, by training on \bookcorefsilver, Longdoc scores 67.5 CoNLL-F1 points on the Animal Farm benchmark, an increase of 31.2 points compared to previous literature~\citep{guo-etal-2024-dual}.
We reserve Appendix~\ref{appendix:more-results} for a more detailed report on the performance and robustness of the comparison systems.

\paragraph{\bookcorefgoldwindows} Results on \bookcorefgoldwindows\ show an interesting finding: both off-the-shelf and fine-tuned systems perform much better in this medium-sized text setting, with \maverickxl\ reaching the best results.
Specifically, \maverickxl\ scores 82.2 CoNLL-F1 points when trained on the \bookcorefsilver\ data, surpassing the second-best model, \dualcache,  by +4.9 points.

We note that this result aligns with previous work, demonstrating that the incremental formulation of \longdoc is particularly robust on full books, whilst Maverick's single forward pass approach is superior on medium-sized text.

\paragraph{Metrics}
Comparing the measures of the full-book setting with the ones obtained with smaller windows unveils a recurring pattern:
traditional coreference scores, i.e., \ceafe\ \bcubed\ and MUC, show discrepancies within each comparison system when evaluating full-book performance.

Interestingly, MUC often measures over 80 F1 points both in the window and book-scale setting, while \bcubed\ and, in particular, \ceafe, often show lower scores on full books, which substantially increase when evaluating model performance on smaller texts.

While many previous works have shown weaknesses and disagreements of these metrics in specific scenarios \citep{moosavi-strube-2016-coreference,borovikova-etal-2022-methodology,durontejedor2023evaluatecoreferenceliterarytexts}, we argue that this new proposed book-scale setting offers fresh ground for improving the study of robustness and expressivity of standard CR metrics.
We reserve Section~\ref{appendix:more-results} in the Appendix to obtain further insights into the behavior of coreference metrics and models in our book-scale scenario.
Specifically, we propose an intermediate setting where full-book predictions are evaluated on smaller windows, and find that the discrepancy between CR metrics is lower in this setting compared to the original \bookcorefgold.
These additional results highlight the need to increase the robustness of CR metrics in the book setting.


\begin{table}[]
\centering

{\small
\begin{tabular}{|l|c|c|}
\hline 

& \multicolumn{1}{c|}{Time (sec.) } 
& \multicolumn{1}{c|}{Memory (GB)} 
\\

\hline

\rowcolor{gray!20}
\multicolumn{3}{|c|}{Off-the-shelf}
\\
BookNLP & 180 & ~6.9 \\

\longdoc & \textbf{~70} & \textbf{~5.8} \\

\dualcache & 460 & 11.6 \\

\maverickxl & 211 & 14.8 \\

\hline
\rowcolor{gray!20}
\multicolumn{3}{|c|}{Trained on \bookcorefsilver}
\\

\longdoc & \textbf{~26} & \textbf{~5.8} \\

\dualcache  & 473 & 11.8 \\

\maverickxl &  183 & 12.8  \\

\hline
\end{tabular}%
}
\caption{Efficiency of each model when running inference on \bookcorefgold. We measure the total execution time in seconds and the maximum GPU memory in gigabytes, marking the best scores in bold. All experiments were run on an NVIDIA RTX-4090.}
\label{tab:efficiency}
\end{table}

\paragraph{Efficiency}
In Table \ref{tab:efficiency} 
we detail the time and memory requirements of the comparison systems on our full book evaluation.
Our results show that \longdoc\ achieves the lowest inference time and memory usage when processing \bookcorefgold.
We also note that \longdoc\ improves its time efficiency after being trained on \bookcorefsilver.
We believe that the model learns to extract fewer, more precise mentions, which impacts its incremental clustering step and reduces processing time to 26 seconds.


\paragraph{Open Challenges of Book-scale CR}
Our results show that \longdoc achieves the best scores on book-scale coreference resolution after training on \bookcorefsilver.

We point out that the score of 67.0 CoNLL-F1 points on \bookcorefgold\ is relatively low compared to the score models attain on other CR benchmarks such as OntoNotes and LitBank, which is around 80 CoNLL-F1 points.
Nevertheless, the results obtained on \bookcorefgoldwindows\ show that models can reach similar scores when dealing with smaller windows of text, attaining 82.2 CoNLL-F1 points. 
This demonstrates that the difference in performance between \bookcorefgold\ and \bookcorefgoldwindows\ is not due solely to the lack of manually annotated training resources.
On the contrary, we argue that this difference highlights the open research challenges of our new book-scale setting, such as i) studying the interpretation of current coreference metrics for full-book evaluation, ii) creating new systems that can fully exploit book annotations, and iii) exploring efficient solutions for adapting current generative and encoder-only models for book-scale processing without incurring exponential computational costs.

\section{Conclusion}
In this paper, we fill the current gap in book-scale coreference resolution by proposing \bookcoref, a novel benchmark with unprecedented book-scale characteristics.
We put forward \bookcorefsilver, a silver training corpus, and \bookcorefgold, a manually annotated dataset for model evaluation.
\bookcorefsilver\ is annotated using the \bookcorefpipeline, a novel procedure that can provide full-text coreference annotations of the characters in a book, starting from the character names.
We carefully validate our automatic approach intrinsically, measuring its effectiveness on our manually annotated dataset, and also extrinsically, showing that long-document systems benefit from training on our silver resource.
Nevertheless, our results on \bookcorefgold show that specifically tailored long-document systems have a notable decrease in performance when evaluated on full books compared to that on smaller texts.

By releasing \bookcoref, we enable the investigation of the new research challenges posed by the book-scale setting, such as developing more accurate systems and studying long-document metrics.

\section{Limitations}
Although we measure and report the high-quality annotations of \bookcorefsilver, we remark that it is created automatically, and therefore may contain errors.
The lack of manually annotated data for training may represent a possible limit in creating more robust book-scale CR systems and should be addressed in future annotation efforts.

Moreover, the coreference annotations we provide are limited to book characters.
This decision is aligned with previous work on literary CR and is strongly motivated from a narrative perspective.
We argue that, as a first foray into truly book-scale Coreference Resolution, this trade-off will still enable future research on this new setting.


We also note that our test set is composed of canonical texts, i.e., books that have been studied extensively and are readily available on the Internet.
Testing with less renowned books would benefit the evaluation capabilities of \bookcorefgold, a direction that should be prioritized by future work.

Furthermore, our implementation of the \bookcorefpipeline\ is currently focused on the English language, potentially limiting its broader applicability.
However, we note that integrating multilingual automatic systems for Character Linking and Coreference Resolution into the \bookcorefpipeline\  would be enough to create multilingual corpora.
We leave this intuition to future work.

Finally, we could not test some of the currently available systems on the book-scale setting.
Our experiments were limited by hardware availability, i.e., a single RTX-4090. 
For this reason, we could not benchmark large generative models as we did with encoder-only systems.

\section*{Acknowledgements}
\begin{center}
\noindent
    \begin{minipage}{0.1\linewidth}
        \begin{center}
            \includegraphics[scale=0.05]{acknowledgements/fair.pdf}
        \end{center}
    \end{minipage}
    \hspace{0.05\linewidth}
    \begin{minipage}{0.60\linewidth}
         We gratefully acknowledge the support of the PNRR MUR project PE0000013-FAIR.
    \end{minipage}
    \hspace{0.03\linewidth}
    \begin{minipage}{0.1\linewidth}
        \begin{center}
            \includegraphics[scale=0.08]{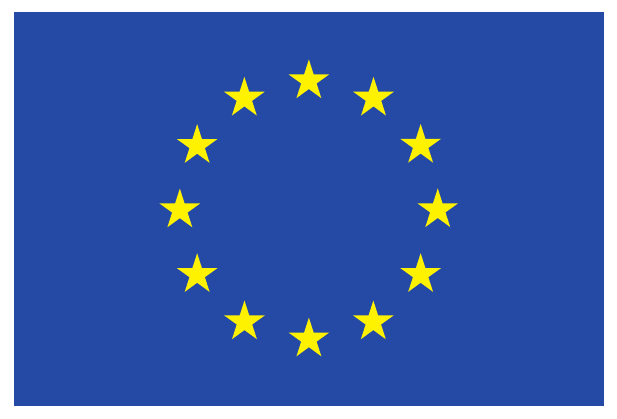}
        \end{center}
    \end{minipage}\\
\end{center}
\vspace{0.2cm}
\noindent Roberto Navigli also gratefully acknowledges the support of the CREATIVE project (CRoss-modal understanding and gEnerATIon of Visual and tExtual content), which is funded by the MUR Progetti di Rilevante Interesse Nazionale programme (PRIN 2020).
This work has been carried out while Giuliano Martinelli was enrolled in the Italian National Doctorate on Artificial Intelligence run by Sapienza University of Rome. 
\bibliography{anthology, custom}

\appendix

\section{Resource Details\label{appendix:liscu-gutenberg}}


The authors of LiSCU~\citep{brahman-etal-2021-characters-tell} do not directly release their data but instead redirect to a GitHub repository to scrape the Internet Archive and obtain the full dataset.\footnote{\small\url{https://github.com/huangmeng123/lit_char_data_wayback}}
We fix broken links, remove some restrictive filters from their codebase that skip minor characters, and run their scripts to save a local copy of LiSCU in a local DB.
We then export a list of 24,985 tuples, each specifying a character name, the author and title of the book it is linked to, and the internet study guide that reports this information.
The total number of books extracted from LiSCU is 1,707, each linked to 14.6 characters on average.

For Wikidata, we derive our own SPARQL query to use three main Wikidata relations:
\textit{characters} (\href{https://www.wikidata.org/wiki/Property:P674}{P674}), which links an item such as a book or a film to characters that appear in it; \textit{author} (\href{https://www.wikidata.org/wiki/Property:P50}{P50}), which links to the author of a specific work; and \textit{title} (\href{https://www.wikidata.org/wiki/Property:P1476}{P1476}), the published name of a literary work or newspaper article or others.
Running this query results in 17,324 distinct (author, title) tuples, each linked on average to 1.54 characters.
This highlights some issues of the Wikidata Knowledge Graph, as the character relation has poor coverage and there is no reliable way to restrict an item identified by an (author, title) tuple to be a work of literary text.

We then attempt to match each (author, title) obtained from the concatenation of LiSCU and Wikidata to a valid book in Project Gutenberg.
We use the \texttt{gutenbergpy} library\footnote{\small\url{https://github.com/raduangelescu/gutenbergpy}}, which includes functions to search through the whole Project Gutenberg book metadata for an exact token match between our (author, title) tuples and the author and title information in Gutenberg.
There is a possibility of multiple (author, title) matches in our character tuples, as there might be some overlap between the sources they are derived from (Wikidata and the three different study guides contained in LiSCU).
We simply select the source which contains the largest number of characters for a given book.
We manually validate the links between the list of characters of a book and its raw text extracted from Gutenberg, as well as manually cleaning each Gutenberg book to remove headings, placeholders, notes, etc.
This leaves us with our base resource: a set of 52 cleaned books obtained from Project Gutenberg, each linked to a list of characters that appear in the book.

We also include in our corpus \textit{Animal Farm} by George Orwell, as contributed by \citet{guo-etal-2024-dual}, resulting in a total number of 53 books.

\paragraph{Licenses}
Books from Project Gutenberg are available in the public domain in the United States, and the authors of LiSCU~\cite{brahman-etal-2021-characters-tell} make their code freely accessible under an MIT license.
The Animal Farm annotation provided by \citet{guo-etal-2024-dual} is available under the permissive BSD 3-clause license.
We are also allowed to make full use of LitBank under the Creative Commons Attribution 4.0 International License.
Qwen2 7B is licensed with Apache 2.0.
Both ReLiK~\cite{orlando-etal-2024-relik} and Maverick~\cite{martinelli-etal-2024-maverick} are licensed with the Attribution-NonCommercial-ShareAlike 4.0 International License and therefore can be used for research purposes.

\section{Entity Linking on LitBank \label{appendix:relik-litbank}}
Our main aim is to create a dataset that can be used to fine-tune ReLiK~\cite{orlando-etal-2024-relik} to perform Character Linking on the narrative text.
To accomplish this, we adapt the LitBank Coreference Resolution dataset based on the following intuition: 
if we are able to assign a character name to each co-referring cluster of mentions associated to a character, we can then re-frame the task as linking each specific mention to its character name.
More practically, the coreference data in LitBank comes with a categorization of entities across six ACE 2005\footnote{\small\url{https://catalog.ldc.upenn.edu/LDC2006T06}} categories (people, facilities, locations, geo-political entities, organizations and vehicles); we therefore only keep coreference clusters where all mentions are of type \textit{people}, including both named entities and common nouns, which are crucial for many downstream applications~\cite{martinelli-etal-2024-cner}.
Moreover, for each mention in a cluster, we are given its manually-validated part-of-speech tag.
We filter out any pronoun mentions from all the clusters, as they are not usually tagged in Entity Linking settings.
We then manually name each remaining cluster, taking inspiration from the most frequently occurring proper nouns in a cluster but also taking clues from the book (e.g., naming a cluster ``Tom Sawyard'' even if the most frequently occurring PROP is ``Tom'').

This leaves us with a complete Character Linking dataset, i.e., an Entity Linking dataset where all the entities are anthropomorphized characters.
The index of possible entities is the set of named coreference clusters of a book and is different for each book.

We then follow the instructions in the ReLiK repository\footnote{\small\url{https://github.com/SapienzaNLP/relik}} to fine-tune the ReLiK Reader on the LitBank training set.
Evaluating on the held-out testing set returned an F1 score of 83.1.

\begin{table}[h]
    \centering
    \begin{tabular}{p{0.9\columnwidth}}
         \hline
         \textbf{Prompt} \\
         \hline
         I will give you an excerpt from a book with a highlighted mention of a character with []. You will need to answer if the assigned character is correct (Yes), or not (No).
         
         Book excerpt: \$book
         
         Does the mention [\$mention] correspond to the character \$character? (Yes/No) \\
         \hline
    \end{tabular}
    \caption{Prompt template for the Cluster Refinement step. The dollar sign (\$) indicates a template variable.}
    \label{tab:qwen2-prompt}
\end{table}

\section{Prompting LLMs for Cluster Refinement\label{appendix:llm}}
We detail our prompt template in Table~\ref{tab:qwen2-prompt}. 
We apply it with a specific mention from a character cluster, together with the 400 words that surround the mention as local context.
We do not provide any in-context examples, as we found the model to perform reliably in a zero-shot setting (line ``LLM filtering'' in Table~\ref{tab:pipeline-ablation}).

A possible edge-case of this setting might be when there is not enough context for the LLM to accurately determine if the mention is correctly linked to a character.
We performed a small experiment by prompting the LLM with an additional clause (i.e., ``If not enough context is provided answer No'') and we measured a negligible difference compared to the values presented in Table~\ref{tab:pipeline-ablation}: with the new prompt, the model performances do not change in precision (+0.0), and decrease in recall and F1 score (-0.2 and -0.1 respectively).
We believe that the equivalence of the two prompting strategies is due to: i) the simplicity of the task, as the considered mentions are explicit (no pronominal mentions are detected by Character Linking) and 42.7\% of them exactly match the name of the character (Pattern Matching baseline in Table~\ref{tab:pipeline-ablation}); and ii) the LLM being able to answer correctly without solely relying on context (i.e., answering correctly because of its parametric memory).

\section{Manual-annotation Details\label{appendix:annotation-guidelines}}

As in the annotation of \textit{Animal Farm}~\citep{guo-etal-2024-dual}, we provide the annotators with a list of characters appearing in each book, and they are tasked with extracting co-referring mentions of each character.
We follow previous works (LitBank, OntoNotes) in selecting as co-referring mentions spans of text that are part of one of the following categories:
proper nouns (for example, \textit{Siddhartha}); common nouns (\textit{the gentleman}); personal pronouns (\textit{he}, \textit{she}); possessive pronouns (\textit{his}, \textit{hers}).

We also decide to mark the maximal extent of a span, resulting in noun phrases such as \textit{``one of the most eminent physicians''}, \textit{``a man whom nobody cared anything about''} or \textit{``the Right Honourable Lady Catherine de Bourgh''}.
This latter example showcases our approach to honorifics, which we include in the maximal span without annotating them separately.

Regarding singleton mentions (i.e., noun phrases that do not co-refer with other mentions), we allow them to be tagged by our annotators, but because the task pertains to identifying co-referring mentions of a list of characters for each book, there are no instances of singleton mentions in \bookcorefgold.

Finally, we also instruct annotators to avoid annotating mentions that can be linked to more than one antecedent mention (referred to as \textit{split-antecedent}), in line with most current Coreference Resolution datasets.

\section{Current Systems Limitations\label{appendix:limitations}}
In this section, we detail the limitations of current systems when applied to this new book-scale setting.

\paragraph{Discriminative models}
Discriminative models formulate the CR task as a classification problem.
They usually adopt the same two-step formulation, in which the model learns to first extract the mentions and then predicts their coreference relations.
Recent models use an underlying encoder-only pre-trained Transformer architecture to encode the input documents. 
This give rise to two main limitations: i) pre-trained Transformers usually have a fixed maximum input length and, ii) their attention mechanism involves a quadratic computational complexity with respect to the input text.
These two characteristics strongly limit their applicability to the proposed book-scale setting, in which input books can reach more than 600.000 tokens.

\paragraph{Sequence-to-Sequence models}
Sequence-to-sequence models have recently proven particularly robust for resolving coreference relations in well-established CR benchmarks\cite{bohnet-etal-2023-coreference, zhang-etal-2023-seq2seq}.
However, the limitations that make the usage of discriminative models difficult in the full-book setting are even more marked when considering Sequence-to-Sequence architectures.
In fact, these models usually rely on very large pre-trained Transformers with several billions of parameters and many works report that their applicability is limited by their computational requirements~\cite{zhang-etal-2023-seq2seq, martinelli-etal-2024-maverick}.
An additional limitation is inherently implied by their application to the coreference task, which requires re-generating the full input text with the addition of special tokens that denote coreference.
In this regard, \citet{bohnet-etal-2023-coreference} reports problems of hallucinations and ambiguous matches of mentions to the input when dealing with the documents in OntoNotes.
For these reasons, we note that the proposed book-scale setting is particularly challenging for generative systems with current formulations, and more advancements are needed to adapt them to this extended-length scenario.

\paragraph{LLMs}


The limitations we raise for large generative Coreference Resolution systems are exacerbated when attempting to apply Large Language Models (LLMs) to this task. 
To the best of our knowledge, there is no clear consensus on how LLMs should be used to extract and resolve mentions of co-referring entities without the pre-identification of mentions or the appearance of hallucinated text.
Previous work also shows that, on the task of Coreference Linking, LLMs do not surpass the performance of smaller, bespoke models on medium-scale settings~\cite{le2024are, porada-cheung-2024-solving}.
Moreover, the scale of our benchmark would constrain us to use only LLMs that have a context window larger than twice the length of the longest document in \bookcorefgold, i.e., more than 300,000 tokens. 
We are excited to see how our benchmark could contribute to convincing progress on these challenges.



\section{Comparison Systems Details\label{appendix:training}}
In this section, we dive into further details regarding the experimental setup presented in Section~\ref{subsection:systems}.

\paragraph{BookNLP} The BookNLP pipeline\footnote{\small\url{https://github.com/booknlp/booknlp}} is a collection of modules based on BERT embeddings, each tasked with carrying out a specific task, such as Entity Recognition, Quotation Attribution, Character Clustering and Coreference Resolution.
The performance of this model on Literary Coreference Resolution tasks and its widespread adoption among research in the Digital Humanities field make it an interesting baseline.
Unfortunately, the library does not provide any option to fine-tune the model for CR, which we speculate might be because of the complex interplay of pipeline components that it relies upon to resolve co-referring mentions at the book scale.
Nevertheless, we adopt it as a baseline in our off-the-shelf setting on both \bookcorefgold\ and \bookcorefgoldwindows.

\paragraph{\longdoc} Introduced by \citet{toshniwal-etal-2020-learning, toshniwal-etal-2021-generalization}, it uses an incremental method for mention clustering, in which mentions are scored against a learned vector representation of previously built clusters.
Specifically, \longdoc\ initializes a ``memory architecture'' that saves previous clusters to compare against.
There are two policies that the memory architecture adopts: ``unbounded'', where all clusters are kept for mention clustering; or ``least-recently used'', where once the maximum number of clusters is reached, the cluster that was used least recently is no longer considered for mention clustering and is thus evicted from memory.
In our experiments, we keep with the default choice encoded in \longdoc\ and adopt the unbounded policy.


We were able to train \longdoc\ from scratch on our \bookcorefsilver\ by using its codebase.\footnote{\small\url{https://github.com/shtoshni/fast-coref}}
Our interventions are minimal: we adapt our dataset to their format and run the training, leaving all hyperparameters unchanged.
Notably, Longdoc uses Longformer as an encoder model, from which the maximum input length is limited to 4096 tokens.
At inference time, longdoc encodes documents in sliding windows and performs the clustering step iteratively over the full input.

\paragraph{\dualcache} \citet{guo-etal-2024-dual} developed this model as a modification of the \longdoc\ memory architecture.
They develop a mechanism that maintains two ``caches'', one controlled by a least-recently used policy (L-cache, or local) and the other by a least-frequently used policy (G-cache, or global).
The model has control over the location of clusters and can swap them between L- and G-cache or remove them entirely depending on the policy value.

The experiments reported by \citet{guo-etal-2024-dual} are based on fine-tuning an existing \longdoc\ model, but they do not release any model weights. 
We therefore replicate their setup by initializing the \dualcache\ encoder from the available \longdoc\ encoder and fine-tuning the whole model on LitBank, adopting the hyperparameters specified in their repository.\footnote{\small\url{https://github.com/QipengGuo/dual-cache-coref/tree/main}}
Having chosen the configuration with sizes of the L- and G-cache set to 25 each, we obtain comparable results: 79.68 CoNLL F1 on LitBank compared to their reported 78.8, and 36.8 CoNLL F1 on Animal Farm against their 36.3.

We then fine-tune this LitBank-trained version of \dualcache\ on \bookcorefsilver, without changing any hyper-parameter.

Notably, \dualcache\ is also based on Longformer.

\paragraph{Maverick}
Maverick~\cite{martinelli-etal-2024-maverick} is an encoder-only model that resolves coreference in a single pass.
We test the \maverickxl architecture, presented in Section \ref{subsection:systems}, which is based on the weights of \textsc{maverick-mes-litbank}.\footnote{\small\url{https://huggingface.co/sapienzanlp/maverick-mes-litbank}}
Specifically, this model is based on DeBERTa-v3-large and pre-trained on LitBank, therefore its maximum input length is around 25,000 tokens.

\section{Coreference Metrics on Long-document \bookcoref \label{appendix:more-results}}
In Section~\ref{section:results}, we show comparison systems reaching higher performances when performing CR on small texts compared to full books.
This finding raises an interesting research question: 
do models under-perform when tested on full books or are metrics penalizing the same predictions when evaluating on the book-scale setting?
In this Section, we provide further experiments to analyze this phenomenon.

\subsection{Experimental Setting}
We benchmark both off-the-shelf and trained systems on the two settings presented in Section~\ref{subsection:exp-design} and propose a third setting to obtain further insights, \bookcorefgoldwindoweval.
We now detail the three proposed settings:
\begin{itemize}
    \item \bookcorefgold, the full-book setting. 
    Models take in input full-books and predict book-level coreference clusters that are scored against book-level gold clusters.
    \item \bookcorefgoldwindows, the split-book setting: 
    Models take in input books split into independent windows of 1500 tokens and produce window-level clusters, which are scored against the respective window-level gold clusters.
    \item \bookcorefgoldwindoweval, an intermediate setting, in which full-book predictions are evaluated in windows.
    Specifically, models take in input full books and predict book-level coreference clusters, as in the full-book setting. 
    Then, performance is computed by considering only the predictions that refer to the specific windows used in \bookcorefgoldwindows, scoring them against the respective window-level gold clusters. 
\end{itemize}
Introducing our third intermediate setting is crucial for our investigation in the behavior of coreference metrics and models in our book-scale scenario.

Comparing the results between \bookcorefgold\ and \bookcorefgoldwindoweval\ is useful to evaluate whether the same predictions are underestimated by traditional coreference metrics.
On the other hand, comparing the outcome of  \bookcorefgoldwindows\ and \bookcorefgoldwindoweval\ is useful to investigate changes in performances when having the full book context.

\begin{table*}[]
\centering
\resizebox{\textwidth}{!}{%
\begin{tabular}{|l|||c|c|c||c||c|c|c||c|||c|c|c||c|||}
\hline 

& \multicolumn{4}{c|||}{\bookcorefgold} 
& \multicolumn{4}{c|||}{\bookcorefgoldwindoweval} 
& \multicolumn{4}{c|||}{\bookcorefgoldwindows} 

\\
           
\hline

& \multicolumn{1}{c|}{MUC}  
& \multicolumn{1}{c|}{\bcubed} 
& \multicolumn{1}{c|}{\ceafe} 
& \multicolumn{1}{c|||}{CoNLL} 

& \multicolumn{1}{c|}{MUC} 
& \multicolumn{1}{c|}{\bcubed} 
& \multicolumn{1}{c|}{\ceafe} 
& \multicolumn{1}{c|||}{CoNLL}

& \multicolumn{1}{c|}{MUC}  
& \multicolumn{1}{c|}{\bcubed} 
& \multicolumn{1}{c|}{\ceafe} 
& \multicolumn{1}{c|||}{CoNLL}

\\ 
\hline
\rowcolor{gray!17}
\multicolumn{13}{|c|||}{Off-the-shelf}
\\

BookNLP & \underline{83.1} & 40.9 & ~2.4 & 42.2 & 81.3 & 54.3 & 23.4 & 53.0 & 81.1 & 52.3 & 18.7 & 50.6\\
\longdoc & 79.9 & \underline{52.1} & \underline{~7.9} & \underline{46.6} & 80.3 & 64.2 & 34.2 & 59.4 & 80.7 & 63.8 & 39.0  & 61.2\\
\dualcache & 82.3 & 41.0 & ~4.2 & 42.5 & \underline{82.6} & \underline{67.4} & \underline{41.3} & \underline{63.7} & 82.9 & 67.8 & 43.9 & 64.8\\
\maverickxl & 81.2 & 35.6 &  ~6.1 & 41.2 & 80.9 & 55.9 & 26.2 & 54.3 & \underline{83.8} & \underline{69.5} & \underline{46.1} & \underline{66.5} \\
\hline
\rowcolor{gray!17}
\multicolumn{13}{|c|||}{Fine-tuned on \bookcorefsilver}
\\
\longdoc & 93.5 & \textbf{62.4} & \textbf{45.3} & \textbf{67.0} & 80.8 & \textbf{75.3} & \textbf{62.2} & \textbf{76.2} & 91.2 & 74.9 & 65.7 & 77.1 \\

\dualcache & 92.5 & 48.0 & 16.9 & 52.5 & 90.7 & 73.5 & 61.4 & 75.2 & 91.2 & 74.5 & 66.2 & 77.3 \\

\maverickxl & \textbf{94.3} & 55.3 & 33.4 & 61.0 & \textbf{91.7} & 72.8 & 55.6 & 73.4 & \textbf{92.7} & \textbf{82.2} & \textbf{71.9} & \textbf{82.2} \\

\hline

\end{tabular}%
}
\caption{Comparison between off-the-shelf models and systems trained on \bookcorefsilver, tested on \bookcorefgold\, \bookcorefgoldwindows and \bookcorefgoldwindoweval. For each system, we report F1 measures of MUC, \bcubed\ and \ceafe, and use their average, CoNLL-F1, as the main evaluation criteria.  We highlight in \textbf{bold} the best measures of trained models, and \underline{underline} best off-the-shelf results.}
\label{tab:full}
\end{table*}

\subsection{Results}

\paragraph{Performance of comparison systems}
In Table \ref{tab:full} we report the results of our comparison systems in the three proposed settings.
Interestingly, results on \bookcorefgoldwindoweval\ are consistently higher compared to the ones measured in \bookcorefgold, indicating that traditional coreference metrics are biased towards lower performances when evaluating full-book predictions.
Furthermore, when evaluated on \bookcorefgoldwindoweval, the score increase is particularly high for Dual-cache, suggesting that this model is locally accurate but lacks full-book consistency.
The same cannot be said for Longdoc, which in the \bookcorefgoldwindoweval\ setting confirms the high performances obtained on full-book predictions.
In general, all the systems perform better when taking in input medium-sized texts, as on \bookcorefgoldwindows, suggesting that the current models are particularly tailored for processing shorter lengths.

\paragraph{Coreference metrics}
In the full book setting, we notice a low agreement between the MUC and \ceafe\ metrics.
In particular, MUC scores are typically very high, whilst \ceafe\ instead assigns a low evaluation to the same predictions.
This gap between these two metrics is less evident on \bookcorefgoldwindoweval, in which the same predictions are evaluated in smaller windows.
Since analyzing this phenomenon is not in the scope of this paper, we leave this research direction to future work.
Nevertheless, we believe that a more detailed study is needed for analyzing metrics behavior in full-book settings, and more attention should be devoted to the interpretation of their scores.

\newpage

\begin{table*}[htbp]
\centering
\begin{tabular}{p{0.24\textwidth}|p{0.74\textwidth}}
\hline\hline
\textbf{Stage} & \textbf{Annotation} \\
\hline\hline
Character Linking (CL) & \corefPurple{Miss Bingley}{Caroline Bingley} sees that \corefRed{her brother}{Mr. Darcy} is in love with you and wants him to marry \corefOlive{Miss Darcy}{Georgiana Darcy}. \\
\hline
CL + LLM Filtering & \corefPurple{Miss Bingley}{Caroline Bingley} sees that her brother is in love with you and wants him to marry \corefOlive{Miss Darcy}{Georgiana Darcy}. \\
\hline
CL + Window Coreference & \corefPurple{Miss Bingley}{Caroline Bingley} sees that \corefRed{\corefPurple{her}{Caroline Bingley} brother}{Mr. Darcy} is in love with \corefBlue{you}{Jane Bennet} and wants \corefRed{him}{Mr. Darcy} to marry \corefOlive{Miss Darcy}{Georgiana Darcy}. \\
CL + LLM Filtering + Window Coreference & \corefPurple{Miss Bingley}{Caroline Bingley} sees that \corefGreen{\corefPurple{her}{Caroline Bingley} brother}{Charles Bingley} is in love with \corefBlue{you}{Jane Bennet} and wants \corefGreen{him}{Charles Bingley} to marry \corefOlive{Miss Darcy}{Georgiana Darcy}.\\
\hline\hline
CL & When , after examining \corefOlive{the mother}{Lady Catherine de Bourgh} , in whose countenance and deportment she soon found some resemblance of \corefBrown{Mr. Darcy}{Mr. Darcy} , she turned her eyes on \corefRed{the daughter}{Charlotte Lucas} , she could almost have joined in \corefPurple{Maria}{Maria Lucas} ’s astonishment at her being so thin and so small . \\
\hline
CL + LLM Filtering & When , after examining the mother , in whose countenance and deportment she soon found some resemblance of Mr. Darcy , she turned her eyes on the daughter , she could almost have joined in \corefPurple{Maria}{Maria Lucas} ’s astonishment at her being so thin and so small . \\
\hline
CL + Window Coreference & When , after examining \corefOlive{the mother}{Lady Catherine de Bourgh} , in whose countenance and deportment \corefBlue{she}{Elizabeth Bennet} soon found some resemblance of \corefBrown{Mr. Darcy}{Mr. Darcy} , \corefBlue{she}{Elizabeth Bennet} turned \corefBlue{her}{Elizabeth Bennet} eyes on \corefRed{the daughter}{Charlotte Lucas} , \corefBlue{she}{Elizabeth Bennet} could almost have joined in \corefPurple{Maria}{Maria Lucas} 's astonishment at \corefRed{her}{Charlotte Lucas} being so thin and so small . \\
CL + LLM Filtering + Window Coreference & When , after examining \corefOlive{the mother}{Lady Catherine de Bourgh} , in whose countenance and deportment \corefBlue{she}{Elizabeth Bennet} soon found some resemblance of \corefBrown{Mr. Darcy}{Mr. Darcy} , \corefBlue{she}{Elizabeth Bennet} turned \corefBlue{her}{Elizabeth Bennet} eyes on \corefGreen{the daughter}{Miss de Bourgh} , \corefBlue{she}{Elizabeth Bennet} could almost have joined in \corefPurple{Maria}{Maria Lucas} 's astonishment at \corefGreen{her}{Miss de Bourgh} being so thin and so small . \\
\hline\hline
\end{tabular}
\caption{Error analysis of our \bookcorefpipeline\ predictions on \textit{Pride and Prejudice} from our test set. The table examines the three main pipeline steps as defined in Table~\ref{tab:pipeline-ablation} (Cluster Initialization, Cluster Refinement, and Cluster Expansion), through two examples.
The first example demonstrates effective LLM Filtering, which correctly removes the mention ``\corefRed{her brother}{Mr. Darcy}'' that should link to `Charles Bingley'. Without this filtering, Cluster Expansion would propagate the error by including nearby pronouns. When applied after LLM Filtering, Cluster Expansion successfully identifies the correct character by linking coreferential mentions across a wider context.
The second example illustrates the pipeline's robustness even when LLM Filtering over-corrects by removing valid mentions like ``\corefBrown{Mr. Darcy}{Mr. Darcy}''. The Cluster Expansion step recovers these filtered mentions while also correcting incorrectly linked mentions from the initial clustering (``\corefRed{the daughter}{Charlotte Lucas}'').
}
\label{table:error-analysis}
\end{table*}

\begin{table*}
    \centering
    {\small
    \begin{tabularx}{0.8\textwidth}{
    l
    l
    >{\raggedright\arraybackslash}X
    >{\centering\arraybackslash}X
    >{\centering\arraybackslash}X 
    }
    \hline
    & Book name & Gutenberg key & Number of tokens & Number of characters \\
    \hline
    & The Pilgrim's Progress & 7088 & ~30610 & 45 \\
    & The Boxcar Children & 42796 & ~30819 & 11 \\
    & Dr. Jekyll and Mr. Hyde & 42 & ~31127 & ~8 \\
    \rowcolor{LimeGreen!20}
    G & Animal Farm* & - & ~36504 & 20 \\
    \rowcolor{LimeGreen!20}
    G & Siddhartha & 2500 & ~47785 & ~9 \\
    & The Awakening & 160 & ~60031 & 26 \\
    & Pudd'nhead Wilson & 102 & ~63061 & 20 \\
    & O Pioneers! & 24 & ~67463 & 12 \\
    & The Hound of the Baskervilles & 2852 & ~70582 & 19 \\
    & The Prince and the Pauper & 1837 & ~81947 & 21 \\
    & Silas Marner & 550 & ~87218 & 33 \\
    & Northanger Abbey & 121 & ~92861 & 17 \\
    & My Ántonia & 19810 & ~95698 & 38 \\
    & Kidnapped & 421 & ~98379 & 21 \\
    & Persuasion & 105 & ~99102 & 20 \\
    & Fathers and Sons & 47935 & ~99902 & 27 \\
    & The Tale of Genji & 66057 & 106868 & 20 \\
    & Pan Tadeusz & 28240 & 136711 & 10 \\
    & Sense and Sensibility & 161 & 142607 & 18 \\
    & The Mayor of Casterbridge & 143 & 143129 & 13 \\
    \rowcolor{LimeGreen!20}
    G & Pride and Prejudice & 1342 & 146495 & 38 \\
    & The Talisman & 1377 & 155678 & 10 \\
    & Moll Flanders & 370 & 159287 & 14 \\
    & In Search of the Castaways & 46597 & 166845 & 10 \\
    & A Tale of Two Cities & 98 & 170076 & 14 \\
    & The Last of the Mohicans & 27681 & 172400 & 12 \\
    & The Return of the Native & 122 & 174304 & 18 \\
    & The Jungle & 140 & 177580 & 46 \\
    & The Way of All Flesh & 2084 & 185562 & 39 \\
    & Emma & 158 & 190921 & 24 \\
    & The Canterbury Tales & 2383 & 205165 & 45 \\
    & Main Street & 543 & 209394 & 84 \\
    & The Red and the Black & 44747 & 222200 & 24 \\
    & The Man in the Iron Mask & 2759 & 222328 & 33 \\
    & Ivanhoe & 82 & 224176 & 70 \\
    & North and South & 4276 & 224604 & 11 \\
    & Little Women & 514 & 229337 & 61 \\
    & Jane Eyre & 1260 & 231978 & 34 \\
    & Uncle Tom's Cabin & 203 & 233220 & 25 \\
    & The Deerslayer & 3285 & 253815 & 15 \\
    & Pamela & 6124 & 267364 & 35 \\
    & The Three Musketeers & 1257 & 292259 & 31 \\
    & The Woman in White & 583 & 294178 & 13 \\
    & Twenty Years After & 1259 & 315692 & 20 \\
    & Middlemarch & 145 & 378961 & 35 \\
    & The Pickwick Papers & 580 & 388962 & 88 \\
    & Our Mutual Friend & 883 & 417160 & 17 \\
    & Little Dorrit & 963 & 418851 & 15 \\
    & The Brothers Karamazov & 28054 & 435055 & 12 \\
    & Bleak House & 1023 & 437143 & 66 \\
    & Don Quixote & 996 & 473968 & 31 \\
    & War and Peace & 2600 & 675324 & 32 \\
    & Les Miserables & 135 & 689400 & 45 \\
    \hline
    \end{tabularx}
    }
    \caption{List of all books that compose \bookcoref, together with their Gutenberg ID, number of tokens and number of characters. With the color green and the letter G we identify books that were manually annotated to create \bookcorefgold. *: \textit{Animal Farm} is included from \citet{guo-etal-2024-dual} with no further changes.}
    \label{table:bookcoref-books-overview}
\end{table*}

\end{document}